\PassOptionsToPackage{unicode,breaklinks=true,colorlinks=true,urlcolor=Blue}{hyperref}
\PassOptionsToPackage{hyphens}{url}
\documentclass[
]{article}
\usepackage{lmodern}
\usepackage{amsmath} % Always load amsmath
\usepackage{ifxetex,ifluatex}
\ifnum 0\ifxetex 1\fi\ifluatex 1\fi=0 % Check if we are NOT using xetex or luatex
  \usepackage{amssymb}
\else
\fi
\ifnum 0\ifxetex 1\fi\ifluatex 1\fi=0 % if pdftex
  \usepackage[T1]{fontenc}
  \usepackage[utf8]{inputenc}
  \usepackage{textcomp} % provide euro and other symbols
\else % if luatex or xetex
  \usepackage{unicode-math}
  \defaultfontfeatures{Scale=MatchLowercase}
  \defaultfontfeatures[\rmfamily]{Ligatures=TeX,Scale=1}
\fi
\IfFileExists{upquote.sty}{\usepackage{upquote}}{}
\IfFileExists{microtype.sty}{% use microtype if available
  \usepackage[]{microtype}
  \UseMicrotypeSet[protrusion]{basicmath} % disable protrusion for tt fonts
}{}
\makeatletter
\@ifundefined{KOMAClassName}{% if non-KOMA class
  \IfFileExists{parskip.sty}{%
    \usepackage{parskip}
  }{% else
    \setlength{\parindent}{0pt}
    \setlength{\parskip}{6pt plus 2pt minus 1pt}}
}{% if KOMA class
  \KOMAoptions{parskip=half}}
\makeatother
\IfFileExists{xurl.sty}{\usepackage{xurl}}{} % add URL line breaks if available
\IfFileExists{bookmark.sty}{\usepackage{bookmark}}{\usepackage{hyperref}}
\hypersetup{
  pdftitle={Fine-tuning on simulated data outperforms prompting for agent tone of voice},
  pdfauthor={Ingo Marquardt, Philippe Brule Restack July 7, 2025},
  hidelinks,
  pdfcreator={LaTeX via pandoc}}
\usepackage[top=1.2in,bottom=1.2in,left=1.4in,right=1.4in]{geometry}
\usepackage{longtable,booktabs}
\usepackage{etoolbox}
\makeatletter
\patchcmd\longtable{\par}{\if@noskipsec\mbox{}\fi\par}{}{}
\makeatother
\IfFileExists{footnotehyper.sty}{\usepackage{footnotehyper}}{\usepackage{footnote}}
\makesavenoteenv{longtable}
\usepackage{graphicx}
\makeatletter
\def\maxwidth{\ifdim\Gin@nat@width>\linewidth\linewidth\else\Gin@nat@width\fi}
\def\maxheight{\ifdim\Gin@nat@height>\textheight\textheight\else\Gin@nat@height\fi}
\makeatother
\setkeys{Gin}{width=\maxwidth,height=\maxheight,keepaspectratio}
\makeatletter
\def\fps@figure{htbp}
\makeatother
\usepackage{etoolbox}
\AtBeginEnvironment{quote}{\sffamily}

\usepackage{tabularx}
\usepackage{adjustbox}
\usepackage[labelfont=bf,font=sf,format=plain]{caption}

\newlength{\cslhangindent}
\newenvironment{cslreferences}%
  {\setlength{\parindent}{0pt}%
  \everypar{\setlength{\hangindent}{\cslhangindent}}\ignorespaces}%
  {\par}

\title{\textbf{Fine-tuning on simulated data outperforms prompting for
agent tone of voice}}

\author{\textbf{Ingo Marquardt, Philippe Brule}\\
Restack\\
\emph{July 7, 2025}}

\author{\textbf{Ingo Marquardt, Philippe Brule}\\
Restack\\
\emph{July 7, 2025}}
\date{}



\makeatletter
\def\@maketitle{%
  \newpage
  \null
  \vskip 2em%
  \begin{center}%
  \let \footnote \thanks
    {\LARGE \@title \par}%
    \vskip 1.5em%
    {\large
      \lineskip .5em%
      \begin{center}
        \@author
      \end{center}%
    }%
    \vskip 1em%
    {\large \@date}%
  \end{center}%
  \par
  \vskip 1.5em}
\makeatother

\begin{document}
\maketitle
\thispagestyle{empty}

\hypertarget{abstract}{%
\subsection{Abstract}\label{abstract}}

Deploying language models (LMs) in customer-facing speech applications
requires conversational fluency and adherence to specific stylistic
guidelines. This can be challenging to achieve reliably using complex
system prompts due to issues like instruction following limitations and
in-context bias. This study investigates the effectiveness of
fine-tuning versus system prompting for aligning LMs with a specific
behavioral target: responding in a natural, conversational tone suitable
for voice interactions. We fine-tuned a small, open-weights model
(\texttt{Llama3.2-1B-Instruct}) using Low-Rank Adaptation (LoRA) on a
synthetically generated dataset derived from Wikipedia. Additionally, we
fine-tuned two closed-source models (\texttt{gpt-4o-mini},
\texttt{gpt-4.1-mini}). Our results demonstrate that fine-tuning
outperformed system prompting, achieving a high percentage of
conversational responses, even when trained on only 100 data samples.
Semantic similarity analysis confirmed that fine-tuning did not degrade
content quality. Interestingly, fine-tuning with 8-bit integer
quantization converged faster towards the target style than using
bfloat16 precision, potentially due to implicit regularization effects.
We conclude that fine-tuning small, open-weights LMs on simulated data
is a highly effective and data-efficient method for instilling specific
stylistic behaviors, offering a preferable alternative to complex system
prompting for practical applications requiring nuanced response styles.

\hypertarget{introduction}{%
\subsection{1. Introduction}\label{introduction}}

Deploying language models (LMs) in real-world, customer-facing speech
applications requires conversational fluency, low latency, and adherence
to specific rules. Besides giving factually correct responses, a
customer-facing LM is required to respond in an appropriate tone of
voice, and to follow application-specific guidelines. In practice, this
is usually achieved by including a list of instructions in the system
prompt. While this approach can work well for demonstrations with a
limited number of system instructions, a growing list of complex prompts
can result in suboptimal instruction following (Wen et al. 2024).

Selecting the right prompt for a given task is far from trivial (Lee,
Kang, and Yoo 2025; Kusano, Akimoto, and Takeoka 2024; Polo et al.
2024). Occasionally, LMs might not even follow a single system
instruction. Language is ambiguous and context dependent, and while the
intended outcome might be perfectly clear to the developer writing the
system prompt, the LM obviously has no knowledge of the developers
implicit goals and the context in which the application is developed. A
popular approach to overcome the ambiguity of language when prompting an
LM is to include one or more example responses in the system prompt.
This approach is known as in-context learning (Brown et al. 2020).

In-context learning has emerged as a powerful technique for aligning LMs
with specific tasks without parameter updates. However, in-context
learning works best for narrow use cases, where a simple input-response
mapping is known beforehands, and the in-context examples can cover the
expected distribution of inputs. In even moderately complex use cases,
in-context learning can suffer from a significant challenge known as
example-choice bias or in-context bias, where models inadvertently learn
spurious correlations from demonstration examples, rather than the
intended task pattern. As an example, consider using an LM for
classifying the sentiment of user reviews (positive, negative, neutral).
If the in-context examples include a review classified as positive for
one product category (e.g.~a toothbrush), and a negative review for
another product category (e.g.~a phone), the model outputs might be
biased (i.e.~have a tendency to classify toothbrush reviews more
positively than phone reviews). While the potential for bias is obvious
in this simple example, in more complex applications it is not obvious
how to avoid example-choice bias. This phenomenon threatens the
reliability of in-context learning applications.

Several studies have investigated example-choice bias. Min et al. (2022)
demonstrated that LMs can fixate on superficial patterns in examples
while disregarding explicit task instructions. Zhao et al. (2021)
identified multiple bias types in in-context learning, including
position bias and majority label bias, highlighting the need for careful
calibration. Lu et al. (2022) revealed that example ordering
significantly impacts performance, suggesting that models are highly
sensitive to demonstration sequencing. Because of such biases, altering
prompts of an LM that is deployed in an AI application can have
unintended and unexpected consequences. Hence, long system prompts that
incorporate diverse examples for in-context learning become difficult to
maintain. Furthermore, longer system prompts increase inference latency
and cost. Therefore, it can be beneficial to finetune LMs to align them
with specific system instructions. With fine-tuning, the LM can be
aligned with multiple, complex, application-specific guidelines. This
can be achieved by using a diverse set of examples of the desired
input-response mapping, without increasing response latency at inference
time.

Here, we tested how to align an LM with a behavioral target, either
through fine-tuning or system prompting. Specifically, we fine-tuned a
small, open-weights language model (\texttt{Llama3.2-1B-Instruct};
Grattafiori et al. (2024)) on simulated data. In addition, we also
fine-tuned two closed-source models (\texttt{gpt-4o-mini},
\texttt{gpt-4.1-mini}) on the same data. We compared how well the
behavioral target can be achieved by system prompting alone and with
fine-tuning. As an example for a realistic, practically relevant
behavioral target, we fine-tuned the LM to respond in a natural,
conversational tone. This style directive is practically relevant when
developing a customer-facing speech assistant. When reading aloud the
responses of popular LMs without enforcing such a style directive, the
user experience is not satisfactory. Reading out text in the style of a
blog post or a Wikipedia article does not feel ``natural'' in a
conversation. We demonstrate that to achieve model responses that are
suitable for verbal communication, it is more effective to finetune the
model on a small dataset, rather than using a system prompt. We expect
that our approach generalizes to behavioral targets other than
conversational tone, and that fine-tuning is a preferable choice over
complex system prompts in practical applications with specific style
directives.

\hypertarget{methodology}{%
\subsection{2. Methodology}\label{methodology}}

\hypertarget{model}{%
\subsubsection{2.1 Model}\label{model}}

The goal of our experiment was to achieve model responses in a natural,
conversational tone. Since this behavioral target is practically
relevant in the context of speech assistants, we extracted the
text-to-text component from a multimodal speech-to-text model.
Specifically, we extracted the \texttt{Llama3.2-1B-Instruct} component
from a multimodal Ultravox model
(\texttt{fixie-ai/ultravox-v0\_5-llama-3\_2-1b}). The Ultravox model
combines a speech-to-text model (the encoder component of
\texttt{whisper-large-v3-turbo}; Radford et al. (2022)) with a
pretrained \texttt{Llama3.2-1B-Instruct} text-to-text model. Activations
flow from the speech-to-text model to the text-to-text model, through
what the authors refer to as a multimodal adapter, consisting of two
linear layers.

We chose to use the \texttt{Llama3.2-1B-Instruct} backbone of the
multimodal Ultravox for our experiment because our fine-tuned model can
be directly inserted into the multimodal Ultravox architecture. Thus,
when paired with an additional text-to-speech model, the Ultravox
architecture with our fine-tuned model is well suited for fluent,
natural conversations.

For comparison, in addition to fine-tuning the
\texttt{Llama3.2-1B-Instruct} model, we fine-tuned two closed weights,
commercial models from OpenAI; specifically
\texttt{gpt-4.1-mini-2025-04-14} and \texttt{gpt-4o-mini-2024-07-18}. We
also included the base version of each model in our comparison (i.e.~the
same models before fine-tuning).

\hypertarget{data}{%
\subsubsection{2.2 Data}\label{data}}

\begin{figure}
\centering
\includegraphics{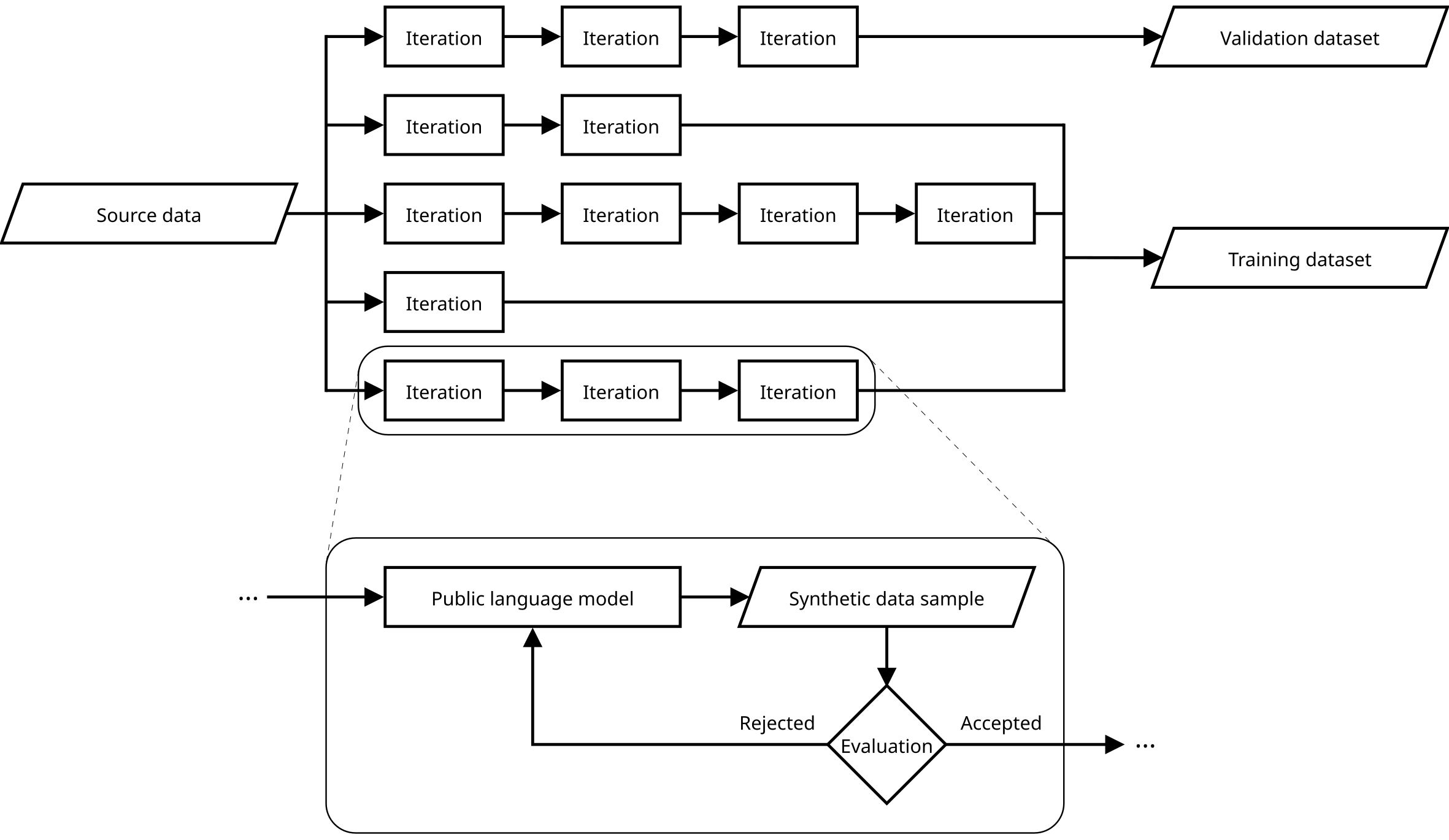}
\caption{The training and validation datasets were generated from
sections of Wikipedia articles using a third party language model
(\texttt{google/gemini-2.0-flash-001}). Scenarios (i.e.~question and
answer pairs) were generated iteratively, i.e.~the third party language
model was prompted repeatedly until the generated answer passed the
criterion for inclusion in the dataset (i.e.~a Flesch reading-ease score
above a specific threshold). For more details on data generation, see
\texttt{A.1\ Data\ generation\ pipeline}.}
\end{figure}

We generated a synthetic dataset consisting of question-answer pairs
based on an open source Wikipedia dataset (Schuhmann 2024). The source
dataset consisted of text sections from articles from the English
Wikipedia. We excluded short text sections (below 700 characters in
length). The remaining text sections had a mean length of 2066
characters. We employed a multi-step data generation pipeline to create
questions and answers from these text sections, using the closed source
\texttt{google/gemini-2.0-flash-001} LM. In a first step, we prompted
the LM to generate an interesting question and the corresponding answer
based on the text section. Secondly, we prompted the LM to rephrase the
answer, in a conversational tone, while keeping the semantic content of
the answer the same.

We used the Flesch reading-ease score to quantify the linguistic style
of the rephrased answers (Kincaid et al. 1975). Flesch reading-ease
score is a text statistic that indicates how difficult a text section is
to understand. A higher score corresponds to easier text. While the name
of the Flesch reading-ease test indicates that it reflects the
\emph{readability} of a text, we found it to be a good indicator of how
natural a text would sound in a verbal conversation. We calculated the
Flesch reading-ease score using the \texttt{textstat} python package
(github.com/textstat/textstat). If the rephrased answer received from
the \texttt{google/gemini-2.0-flash-001} LM achieved a Flesch
reading-ease score equal to or greater than 75, we added the question
and the rephrased answer to the dataset. Conversely, if the reading-ease
score of the rephrased answer was still below 75, we prompted the LM up
to three times to further simplify the linguistic style of the answer.
If the answer did not pass the reading-ease threshold on the third
attempt, the sample was discarded. See Table 1 for an example.

\begin{longtable}[]{@{}ll@{}}
\caption{Representative training sample. The question was generated with
the \texttt{google/gemini-2.0-flash-001} language model based on a
dataset of Wikipedia articles. This particular sample is from the
Wikipedia article \url{https://en.wikipedia.org/wiki/Willow_Plunge}. The
original answer was generated in the first step, together with the
question. Subsequently, the simplified answer was generated in a
separate step. Only simplified answers with a Flesch reading-ease score
equal to or greater than 75 were included in the training and validation
datasets. This particular simplified answer has a Flesch reading-ease
score of 83. See Appendix (\texttt{A.5\ System\ prompts}) for the system
prompts used for these examples.}\tabularnewline
\toprule
\endhead
\begin{minipage}[t]{0.22\columnwidth}\raggedright
Question\strut
\end{minipage} & \begin{minipage}[t]{0.72\columnwidth}\raggedright
Claiborne H. Kinnard Jr.~owned a swimming pool that was later shut down.
What were the reasons for the pool's closure and what remains at the
site today?\strut
\end{minipage}\tabularnewline
\begin{minipage}[t]{0.22\columnwidth}\raggedright
Answer\strut
\end{minipage} & \begin{minipage}[t]{0.72\columnwidth}\raggedright
The pool was shut down in 1967 after the death of Claiborne H. Kinnard
Jr.~in 1966. His widow, Ruth Kinnard, made the decision to close the
pool, after which it was filled in and the land was sold. Today, a
historical marker funded by the Franklin Rotary Club marks the location
of the former swimming pool.\strut
\end{minipage}\tabularnewline
\begin{minipage}[t]{0.22\columnwidth}\raggedright
Simplified answer\strut
\end{minipage} & \begin{minipage}[t]{0.72\columnwidth}\raggedright
Well, the pool closed down in 1967, a year after Claiborne H. Kinnard
Jr.~passed away in 1966. His wife, Ruth Kinnard, decided to shut it
down, so they filled it in and sold the land. Now, all that's left at
the spot is a historical marker that the Franklin Rotary Club paid
for.\strut
\end{minipage}\tabularnewline
\bottomrule
\end{longtable}

The rationale behind basing our synthetic dataset on text sections from
Wikipedia articles was to obtain training data that is domain-agnostic.
The source data was shuffled randomly, so the Wikipedia text sections
covered a diverse variety of subjects. The behavioral target of our
fine-tuning experiment was the linguistic style of the model responses,
not the content. By randomly sampling text sections from Wikipedia
articles, we aimed to obtain a dataset that is not biased towards
particular topics.

Moreover, to ensure a semantically diverse set of questions, and to
avoid duplicates, we compared newly generated questions with those
questions that had already been added to the dataset. We computed the
text embedding vector of each question with an encoder model based on
ModernBERT (\texttt{Alibaba-NLP/gte-modernbert-base}; Li et al. (2023)).
If the embedding vector of a newly generated question had a cosine
similarity of greater than 0.8 with any of the existing question
embeddings, the new question was rejected.

We generated a total of 10,000 question-answer pairs. Of those 10,000
samples, 1000 were assigned to a validation set that was never used
during training. To test the number of samples necessary for
fine-tuning, we created four training subsets consisting of 100, 1000,
5000, and 9000 samples. The dataset is available at
\url{https://huggingface.co/datasets/restack/conversational-question-answer-wikipedia-v1.0}.

\hypertarget{training}{%
\subsubsection{2.3 Training}\label{training}}

The open-weights \texttt{Llama3.2-1B-Instruct} model was trained with
supervised fine-tuning using the Low-Rank Adaptation (LoRA)
implementation from Hugging Face (Hu et al. 2021). The LoRA adapters
targeted the key, query, and value projection layers of the Llama model.
In separate experimental conditions, we trained on each of the four
training subsets (consisting of 100, 1000, 5000, and 9000 samples). In
each case, we trained for 5 epochs with a batch size of 16, and two
gradient accumulation steps, resulting in an effective batch size of 32.
We used the AdamW optimizer (Loshchilov and Hutter 2019) and a cosine
annealing schedule. See Appendix
\texttt{A.3\ Fine-tuning\ hyperparameters\ open-weights} for more
experimental details.

For comparison with the open-weights \texttt{Llama3.2-1B-Instruct}
model, we also fine-tuned two closed-weights models from OpenAI. While
the configuration options in OpenAI's commercial finetuning offering are
limited compared with finetuning open-weights models from the Hugging
Face ecosystem, we tried to match all parameters as closely as possible.
See Appendix \texttt{A.4\ Fine-tuning\ hyperparameters\ closed-weights}
for details.

\hypertarget{evaluation-metrics}{%
\subsubsection{2.4 Evaluation Metrics}\label{evaluation-metrics}}

The goal of our experiment was to test how well the LMs could be aligned
with a specific behavioral target, either through fine-tuning or a
system prompt. Specifically, the behavioral target was to respond in a
natural, conversational tone. Our main outcome variable was the Flesch
reading-ease score (Kincaid et al. 1975), a text statistic that
indicates how difficult a text section is to understand. The easier a
text section, the higher the score. Even though the name of the Flesch
reading-ease test focuses on the \emph{readability} of a text, it is a
good indicator of how natural a text would sound in a verbal
conversation. We used the \texttt{textstat} python package to calculate
the Flesch reading-ease score (github.com/textstat/textstat). See
Appendix (\texttt{A.2\ Example\ model\ responses}, Table 2) for example
model responses with a low / high Flesch reading-ease score.

During fine-tuning, LMs can display deteriorating performance, such as
catastrophic forgetting, where a pretrained language model loses
previously acquired knowledge (Luo et al. 2025). Whereas the Flesch
reading-ease score provides a metric for the linguistic style of a model
response, it does not measure its semantic content. Thus, in order to
ensure that the semantic quality of the model responses does not
deteriorate during fine-tuning, we estimated the semantic similarity
between the predicted model response and the expected model response
during validation. For each sample in the validation set, we computed
the text embedding vectors of the expected answer and of the answer
generated by the fine-tuned model. The encoder model we used was a
ModernBERT variant (\texttt{Alibaba-NLP/gte-modernbert-base}; Li et al.
(2023); Warner et al. (2024)). We quantified the semantic similarity
between the generated answer and the expected answer using the cosine
similarity between the two embedding vectors.

\hypertarget{results}{%
\subsection{3. Results}\label{results}}

During fine-tuning, the open-weights LM quickly achieved the behavioral
target. In other words, the model quickly learned to respond in a
natural, conversational tone. In contrast, prompting the base model
(without fine-tuning) to respond in a natural, conversational tone did
not result in similarly conversational model responses.

To quantify the degree to which the models complied with the behavioral
target, we estimated how conversational model responses were, using the
Flesch reading-ease score (see 3.4 Evaluation Metrics). We defined a
target Flesch reading-ease score of greater than or equal to 60. A model
response that reached or exceeded a score of 60 was deemed sufficiently
conversational. Figure 2 shows the percentage of model responses (out of
all validation samples) that passed the readability score threshold.

\begin{figure}
\centering
\includegraphics{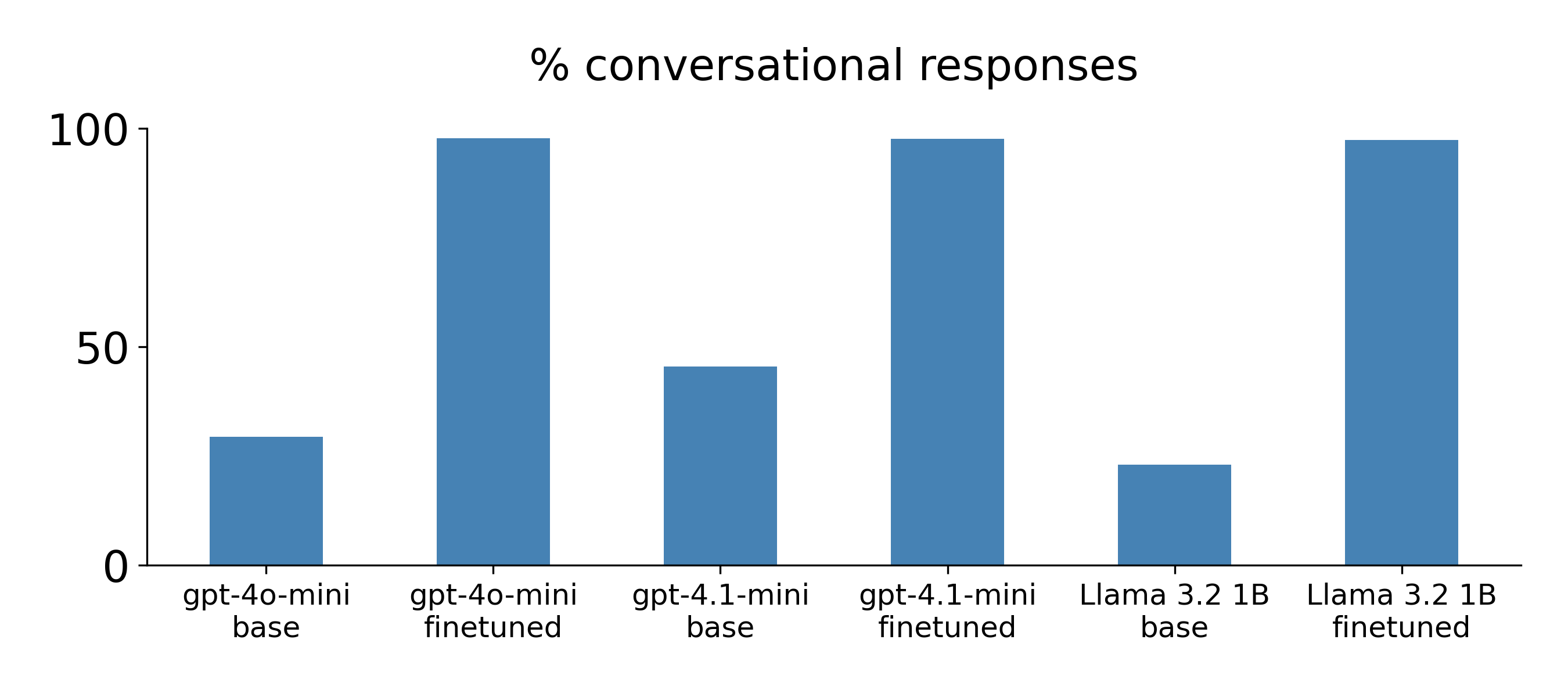}
\caption{Percentage of conversational responses. A model response was
classified as conversational if it reached a Flesch reading-ease score
of at least 60. The bars represent the percentage of model responses
(out of all validation samples) that exceeded this threshold. The
fine-tuned models achieved a much higher percentage of conversational
responses than the base models. The Llama model depicted here was
fine-tuned for 5 epochs on the entire training dataset (9000
question-answer pairs, r = 64, base model loaded in 8 bit integer
precision, learning rate = 2e-4). Likewise, the fine-tuned closed-source
models (\texttt{gpt-4o-mini} and \texttt{gpt-4.1-mini}) were fine-tuned
for 5 epochs on the entire training dataset (9000 question-answer pairs,
see Appendix \texttt{A.4\ Fine-tuning\ hyperparameters\ closed-weights}
for more details). See Appendix (\texttt{A.6\ Detailed\ results}, Table
4) for the corresponding numeric values.}
\end{figure}

Figure 3 shows the percentage of model responses (out of all validation
samples) that passed the readability score threshold, as a function of
training dataset size. With appropriate hyperparameter settings, the
fine-tuned models already reached more than 90\% conversational
responses with only 100 training samples. Larger training dataset sizes
showed diminishing returns, especially for more than 1000 training
samples. For comparison, we also included a larger
\texttt{Llama3.1-8B-Instruct} model fine-tuned with the same
hyperparameters as the most successful \texttt{Llama3.2-1B-Instruct}
model. Model convergence with respect to the behavioral target was very
similar for these two models.

\begin{figure}
\centering
\includegraphics{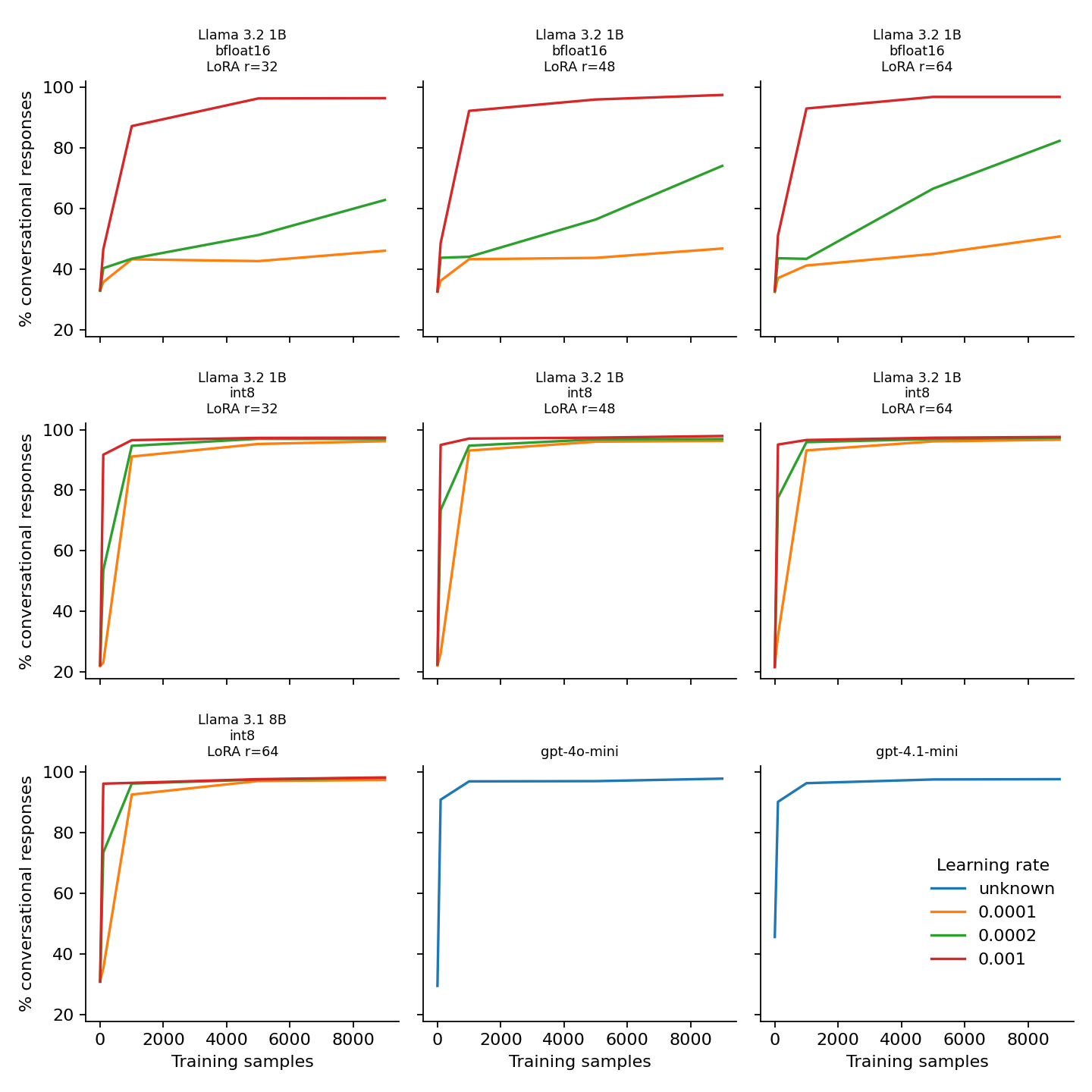}
\caption{Percentage of conversational responses as a function of
training dataset size. We defined a conversational response as a
response that reached a Flesch reading-ease score of at least 60. The
values depicted here represent the percentage of model responses that
exceeded this threshold out of all samples in the validation set. Each
data point corresponds to an experimental condition. For example, the
red line in the top left graph represents five experimental conditions
in which the open-weights \texttt{Llama3.2-1B-Instruct} model was fined
with a specific set of hyperparameters (base model represented in
bfloat16 precision, r = 32, learning rate = 0.001). Each point of the
red line corresponds to a separate experimental run in which a model
with these hyperparameters was fine-tuned on 100, 1000, 5000, or 9000
training samples. In each case, the model was fine-tuned for 5 epochs on
the respective number of training samples. The leftmost point (at
training samples = 0) corresponds to the base model, i.e.~the same model
without fine-tuning. The purpose of this experiment was to investigate
how many training samples are needed for the fine-tuned model to achieve
the behavioral target (i.e.~responding in a conversational tone). The
base models (at training samples = 0) consistently failed to respond in
a conversational tone, even though the system prompt contained
corresponding instructions. Even with just 100 training samples, the
percentage of conversational responses increased substantially, to over
90\% for most model configurations. See Appendix
(\texttt{A.6\ Detailed\ results}, Table 5) for the corresponding numeric
values.}
\end{figure}

We calculated the semantic similarity between the generated model
response and the expected model response to ensure that the quality of
the model predictions did not deteriorate during fine-tuning (for
example due to catastrophic forgetting). As can be seen in Figure 4, we
did not observe deteriorating model predictions under any fine-tuning
condition.

\begin{figure}
\centering
\includegraphics{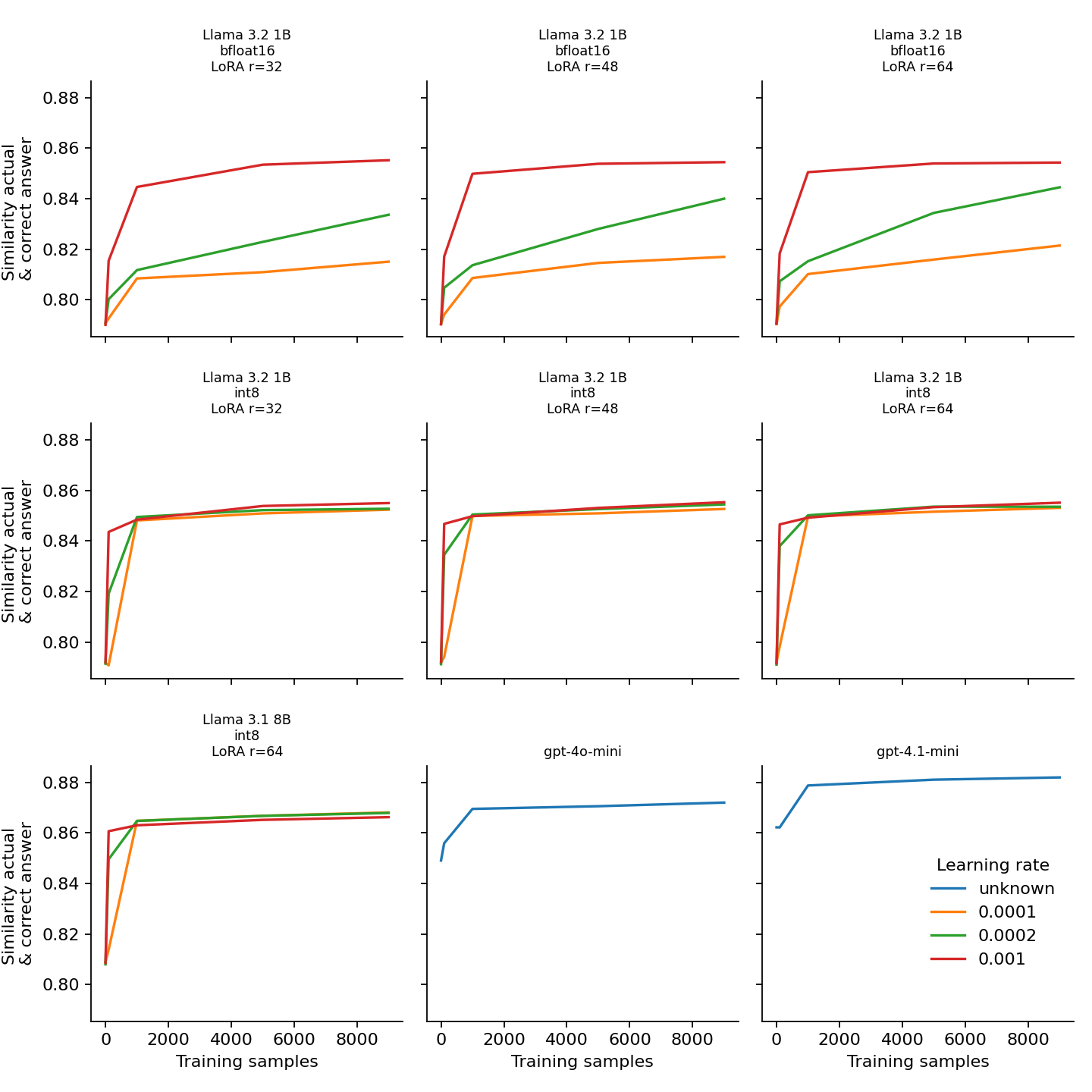}
\caption{Semantic similarity between generated model responses and
expected model responses. Embedding vectors were calculated with a
ModernBERT encoder model (\texttt{Alibaba-NLP/gte-modernbert-base}; Li
et al. (2023); Warner et al. (2024)). We then computed the cosine
similarity between the embedding vectors of the generated and expected
model responses. Under all fine-tuning conditions, the semantic
similarity between the predictions and the expected responses increased
relative to the base model. Hence, there is no evidence for
deteriorating model responses. The \texttt{Llama3.2-1B-Instruct} model
is a very small model. For comparison, we also included a slightly
larger \texttt{Llama3.1-8B-Instruct} model, fine-tuned with the same
hyperparameters as the most successful \texttt{Llama3.2-1B-Instruct}
model. This 8 billion parameter model achieved slightly higher semantic
similarities than the 1 billion parameter model. The presumably much
larger closed-source gpt-4.1 model achieved even higher semantic
similarities. We assume that larger models are more capable at providing
semantically correct responses, therefore achieving a higher score on
this metric. Each data point corresponds to an experimental condition;
see Figure 3 for details. See Appendix (\texttt{A.6\ Detailed\ results},
Table 5) for the corresponding numeric values.}
\end{figure}

\clearpage

\hypertarget{discussion}{%
\subsection{4. Discussion}\label{discussion}}

Our research involved fine-tuning a compact, publicly available language
model to produce responses with a casual, dialogue-like style. This
conversational approach has practical applications in the development of
voice-activated AI systems, particularly those interacting directly with
customers. Notably, our fine-tuned open-source model outperformed a
proprietary commercial model that used system prompting, even when
training with only 100 examples.

Although our experiment centered on creating a conversational tone, we
believe this methodology can be effectively applied to various stylistic
guidelines. Earlier studies have demonstrated that fine-tuning
open-source models works well across numerous tasks (Han et al. 2024; Hu
et al. 2021; Xu et al. 2023). Based on our findings, we suggest that for
practical applications requiring specific stylistic directions,
fine-tuning offers superior results compared to elaborate system
prompting techniques.

Interestingly, during fine-tuning, model convergence towards the
behavioral target was faster (i.e.~required fewer training samples) when
the \texttt{Llama3.2-1B-Instruct} model was loaded with 8-bit integer
quantization (int8), compared with bfloat16 precision (see first and
second row in Figure 3). In the bfloat16 condition, all model weights
were represented in bfloat16 precision. In contrast, in the int8
condition, only trainable parameters (LoRA adapter parameters and the
language modeling head) were represented in bfloat16 precision, whereas
the other model parameters were loaded as 8-bit integers.

Intuitively, higher numeric precision (i.e.~no integer quantization)
should facilitate learning. One possible explanation for why 8-bit
integer quantization actually resulted in better convergence towards the
behavioral target is that the quantization might act as a form of
implicit regularization, preventing overfitting on the small dataset.
Quantization inherently involves mapping values from a higher-precision
representation (like bfloat16 or float32) to a lower-precision one (like
int8; Dettmers et al. (2022)). This process inevitably introduces some
level of approximation error, because the lower-precision format cannot
represent the original values perfectly. While quantization techniques
aim to minimize this error to preserve model accuracy, the residual
noise might function as a form of implicit regularization (Bondarenko,
Chiaro, and Nagel 2024).

In the context of LoRA fine-tuning, the adapters learn based on
gradients derived from the base model's outputs. If the base model is
int8 quantized, the activations passed forward (and thus influencing the
gradients for the adapters) are inherently less precise than in the
bfloat16 case. We hypothesize in our experiment, integer quantization
could have acted as a regularizer, similar to dropout.

Another possible explanation for why int8 quantization of the base model
resulted in better fine-tuning results could be that the added noise
from the quantization helped the optimization process escape local
minima (Bo and Wang 2024). The observed beneficial effect of
quantization was strongest at the lowest learning rate (compare the
orange line in the upper \& middle row in Figure 3). Introducing
quantization-induced noise into the optimization process might have
helped the optimizer escape shallow local minima that could otherwise
trap a full-precision model, especially at low learning rates.

\hypertarget{limitations-and-future-work}{%
\subsection{5. Limitations and Future
Work}\label{limitations-and-future-work}}

Our results indicate that fine-tuning is an effective method for
achieving a specific stylistic target like conversational tone. We would
like to highlight two aspects related to our findings that warrant
further investigation.

Firstly, our observation that 8-bit integer (int8) quantization of the
base model led to faster convergence towards the target style compared
to bfloat16 precision is intriguing, but our proposed
explanations---potential implicit regularization or noise aiding
optimization---remain speculative at this stage. The current work did
not include experiments specifically designed to isolate and confirm
these mechanisms. Future research should conduct targeted experiments to
systematically investigate the interplay between quantization techniques
(like int8) and parameter-efficient fine-tuning methods (like LoRA),
analyzing gradient dynamics and loss landscapes across different
precision levels to provide a more conclusive understanding of this
effect.

Secondly, our primary claim rests on a single, albeit practically
relevant, behavioral target: achieving a conversational tone. While we
hypothesize that fine-tuning generally outperforms complex system
prompting for enforcing diverse style directives, the generalizability
of this conclusion warrants further investigation. Follow-up studies
should conduct systematic comparisons across a broader range of
stylistic targets (e.g., formal tone, specific persona adherence,
complex formatting rules) and potentially different model architectures
and sizes. Such research would help establish the conditions under which
fine-tuning offers the most significant advantages over sophisticated
prompting strategies for controlling LM behavior.

\hypertarget{conclusion}{%
\subsection{6. Conclusion}\label{conclusion}}

We fine-tuned a small, open-weights language model and two
closed-weights models to respond in a natural, conversational tone. A
conversational tone of voice is a style directive that is practically
relevant when developing voice based AI applications, such as a
customer-facing speech assistant. Fine-tuning outperformed system
prompting, even when a very small dataset containing just 100 samples
was used for fine-tuning.

While we have focused on one specific behavioral target (a
conversational tone of voice), we fully expect that our method
generalizes to different style directives. In fact, previous research
provides ample evidence for the feasibility of fine-tuning open-weights
models for a variety of tasks (Han et al. 2024; Hu et al. 2021; Xu et
al. 2023). We argue that fine-tuning is a preferable choice over complex
system prompts in practical applications with specific style directives.

However, the major challenge faced when working with fine-tuned
open-weights models is their deployment. Thanks to the active Hugging
Face software ecosystem, running machine learning experiments has become
easier than ever. In contrast, deploying custom machine learning models
in production is still a considerable challenge. Restack, the sponsor of
this research, offers a backend framework for reliable and scalable
deployment of custom machine learning models. The strong performance of
fine-tuned open-weights models on specific behavioral targets, paired
with the reliability and scalability of Restack's framework for
deploying and managing custom models, provide a compelling case for
purpose-built, application specific machine learning solutions.
Moreover, the Restack framework is ideally suited for the deployment of
AI workflows in general, and in the context of this study, for
generating synthetic datasets (see \texttt{2.2\ Data}). Restack is
committed to provide solutions covering the entire ML development
lifecycle, from data generation, over model training and evaluation, to
model deployment.

\clearpage

\hypertarget{appendix}{%
\section{Appendix}\label{appendix}}

\hypertarget{a.1-data-generation-pipeline}{%
\subsection{A.1 Data generation
pipeline}\label{a.1-data-generation-pipeline}}

\begin{figure}
\centering
\includegraphics{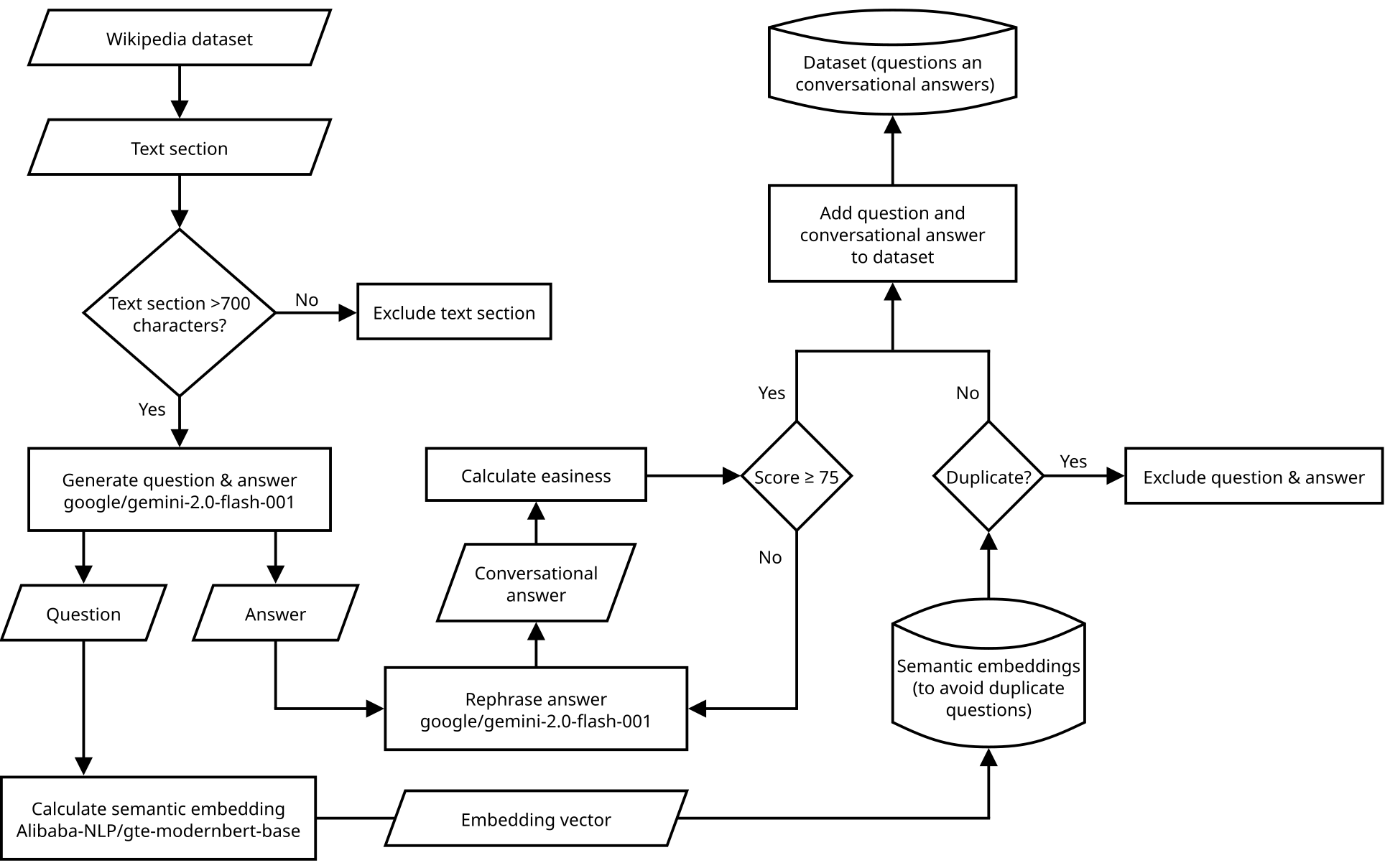}
\caption{Data generation pipeline. Simulated question \& answer pairs
were generated based on a dataset of Wikipedia articles, using a third
party language model (\texttt{google/gemini-2.0-flash-001}). The initial
answer was then rephrased to make it more conversational, using the same
third party model. Only samples with an answer that passed a specific
`easiness' threshold were included. Semantic embeddings of the questions
were used to avoid duplicate or very similar questions, because we aimed
to generate a dataset spanning a diverse range of topics.}
\end{figure}

\clearpage

\hypertarget{a.2-example-model-responses}{%
\subsection{A.2 Example model
responses}\label{a.2-example-model-responses}}

\begin{longtable}[]{@{}lll@{}}
\caption{Example model responses to the user prompt ``In the context of
software engineering, what is the advantage of using a CI / CD
pipeline?''. The fine-tuned models gave more conversational responses.
The \texttt{Llama3.2-1B-Instruct} base model gave a very verbose
response that has been truncated. See Appendix
(\texttt{A.5\ System\ prompts}) for the system prompts used. The Llama
model used here was fine-tuned for 5 epochs on the entire training
dataset (9000 question-answer pairs, r = 64, base model loaded in 8 bit
integer precision, learning rate = 2e-4). Likewise, the fine-tuned
versions of the closed source models (\texttt{gpt-4o-mini} and
\texttt{gpt-4.1-mini}) were fine-tuned for 5 epochs on the entire
training dataset (9000 question-answer pairs, see Appendix
\texttt{A.4\ Fine-tuning\ hyperparameters\ closed-weights} for more
details).}\tabularnewline
\toprule
\begin{minipage}[b]{0.28\columnwidth}\raggedright
Model\strut
\end{minipage} & \begin{minipage}[b]{0.53\columnwidth}\raggedright
Model response\strut
\end{minipage} & \begin{minipage}[b]{0.10\columnwidth}\raggedright
Flesch score\strut
\end{minipage}\tabularnewline
\midrule
\endfirsthead
\toprule
\begin{minipage}[b]{0.28\columnwidth}\raggedright
Model\strut
\end{minipage} & \begin{minipage}[b]{0.53\columnwidth}\raggedright
Model response\strut
\end{minipage} & \begin{minipage}[b]{0.10\columnwidth}\raggedright
Flesch score\strut
\end{minipage}\tabularnewline
\midrule
\endhead
\begin{minipage}[t]{0.28\columnwidth}\raggedright
Llama3.2-1B-Instruct base\strut
\end{minipage} & \begin{minipage}[t]{0.53\columnwidth}\raggedright
So, a CI/CD pipeline is a workflow that automates the build, test, and
deployment of software. And in the context of software engineering, it's
a game-changer. The main advantage of using a CI/CD pipeline is that it
enables continuous integration and continuous deployment (CI/CD). This
means that your code is built, tested, and deployed automatically,
without any manual intervention. Here's what happens: 1. Automated
builds: Your code is compiled, tested, and built automatically, using
tools like Jenkins, Travis CI, or CircleCI. {[}\ldots{]}\strut
\end{minipage} & \begin{minipage}[t]{0.10\columnwidth}\raggedright
34\strut
\end{minipage}\tabularnewline
\begin{minipage}[t]{0.28\columnwidth}\raggedright
Llama3.2-1B-Instruct finetuned\strut
\end{minipage} & \begin{minipage}[t]{0.53\columnwidth}\raggedright
So, a CI / CD pipeline is a way to automate your build and test process.
Think of it like this: you write code, then run it through a test suite.
If everything checks out, you deploy it to the cloud. But if something
goes wrong, the pipeline stops and tells you what's wrong, so you can
fix it fast.\strut
\end{minipage} & \begin{minipage}[t]{0.10\columnwidth}\raggedright
90\strut
\end{minipage}\tabularnewline
\begin{minipage}[t]{0.28\columnwidth}\raggedright
gpt-4o-mini-2024-07-18 base\strut
\end{minipage} & \begin{minipage}[t]{0.53\columnwidth}\raggedright
A CI/CD pipeline automates the process of integrating code changes and
deploying them. This leads to faster development, fewer bugs, and
quicker feedback. It helps teams deliver updates more reliably and
efficiently, improving overall software quality.\strut
\end{minipage} & \begin{minipage}[t]{0.10\columnwidth}\raggedright
34\strut
\end{minipage}\tabularnewline
\begin{minipage}[t]{0.28\columnwidth}\raggedright
gpt-4o-mini finetuned\strut
\end{minipage} & \begin{minipage}[t]{0.53\columnwidth}\raggedright
So, think of it this way: a CI/CD pipeline just automates the boring
stuff. That means everyone on the team can get their code out there
faster and with fewer mistakes. And because things happen quicker and
bugs get caught sooner, the whole team ends up working way more
efficiently.\strut
\end{minipage} & \begin{minipage}[t]{0.10\columnwidth}\raggedright
75\strut
\end{minipage}\tabularnewline
\begin{minipage}[t]{0.28\columnwidth}\raggedright
gpt-4.1-mini-2025-04-14 base\strut
\end{minipage} & \begin{minipage}[t]{0.53\columnwidth}\raggedright
A CI/CD pipeline helps software teams deliver code faster and more
reliably. It automates testing and deployment, catching errors early and
reducing manual work. This leads to quicker updates and better software
quality.\strut
\end{minipage} & \begin{minipage}[t]{0.10\columnwidth}\raggedright
43\strut
\end{minipage}\tabularnewline
\begin{minipage}[t]{0.28\columnwidth}\raggedright
gpt-4.1-mini finetuned\strut
\end{minipage} & \begin{minipage}[t]{0.53\columnwidth}\raggedright
Using a CI/CD pipeline is pretty handy because it lets you add new
features really quickly. Plus, it makes sure your code is checked and
tested automatically all the time. And if there's a problem with a new
version, it's easy to switch back to a working one fast.\strut
\end{minipage} & \begin{minipage}[t]{0.10\columnwidth}\raggedright
75\strut
\end{minipage}\tabularnewline
\bottomrule
\end{longtable}

\hypertarget{a.3-fine-tuning-hyperparameters-open-weights}{%
\subsection{A.3 Fine-tuning hyperparameters
open-weights}\label{a.3-fine-tuning-hyperparameters-open-weights}}

We employed the Transformers / PEFT / LoRA implementations from Hugging
Face. The below table contains the hyperparameters used for the
supervised fine-tuning of the open-weights \texttt{Llama3.2-1B-Instruct}
model.

\begin{longtable}[]{@{}ll@{}}
\caption{Hyperparameters for LoRA fine-tuning of
\texttt{Llama3.2-1B-Instruct}. We iterated over all combinations of the
below hyperparameters.}\tabularnewline
\toprule
\begin{minipage}[b]{0.41\columnwidth}\raggedright
Parameter\strut
\end{minipage} & \begin{minipage}[b]{0.41\columnwidth}\raggedright
Set of values\strut
\end{minipage}\tabularnewline
\midrule
\endfirsthead
\toprule
\begin{minipage}[b]{0.41\columnwidth}\raggedright
Parameter\strut
\end{minipage} & \begin{minipage}[b]{0.41\columnwidth}\raggedright
Set of values\strut
\end{minipage}\tabularnewline
\midrule
\endhead
\begin{minipage}[t]{0.41\columnwidth}\raggedright
Number of epochs\strut
\end{minipage} & \begin{minipage}[t]{0.41\columnwidth}\raggedright
5\strut
\end{minipage}\tabularnewline
\begin{minipage}[t]{0.41\columnwidth}\raggedright
Batch size\strut
\end{minipage} & \begin{minipage}[t]{0.41\columnwidth}\raggedright
16\strut
\end{minipage}\tabularnewline
\begin{minipage}[t]{0.41\columnwidth}\raggedright
Gradient accumulation steps\strut
\end{minipage} & \begin{minipage}[t]{0.41\columnwidth}\raggedright
2\strut
\end{minipage}\tabularnewline
\begin{minipage}[t]{0.41\columnwidth}\raggedright
Effective batch size\strut
\end{minipage} & \begin{minipage}[t]{0.41\columnwidth}\raggedright
32\strut
\end{minipage}\tabularnewline
\begin{minipage}[t]{0.41\columnwidth}\raggedright
Number of training samples\strut
\end{minipage} & \begin{minipage}[t]{0.41\columnwidth}\raggedright
100, 1000, 5000, 9000\strut
\end{minipage}\tabularnewline
\begin{minipage}[t]{0.41\columnwidth}\raggedright
LoRA r \& alpha\textsuperscript{1}\strut
\end{minipage} & \begin{minipage}[t]{0.41\columnwidth}\raggedright
32, 48, 64\strut
\end{minipage}\tabularnewline
\begin{minipage}[t]{0.41\columnwidth}\raggedright
LoRA dropout\strut
\end{minipage} & \begin{minipage}[t]{0.41\columnwidth}\raggedright
0.2\strut
\end{minipage}\tabularnewline
\begin{minipage}[t]{0.41\columnwidth}\raggedright
Precision base model\strut
\end{minipage} & \begin{minipage}[t]{0.41\columnwidth}\raggedright
bfloat16, int8\textsuperscript{2}\strut
\end{minipage}\tabularnewline
\begin{minipage}[t]{0.41\columnwidth}\raggedright
Precision LoRA adapters\strut
\end{minipage} & \begin{minipage}[t]{0.41\columnwidth}\raggedright
bfloat16\strut
\end{minipage}\tabularnewline
\begin{minipage}[t]{0.41\columnwidth}\raggedright
Train bias\strut
\end{minipage} & \begin{minipage}[t]{0.41\columnwidth}\raggedright
Only LoRA adapter biases\strut
\end{minipage}\tabularnewline
\begin{minipage}[t]{0.41\columnwidth}\raggedright
Optimizer\strut
\end{minipage} & \begin{minipage}[t]{0.41\columnwidth}\raggedright
AdamW\strut
\end{minipage}\tabularnewline
\begin{minipage}[t]{0.41\columnwidth}\raggedright
Learning rate (LR) schedular\strut
\end{minipage} & \begin{minipage}[t]{0.41\columnwidth}\raggedright
OneCycleLR\strut
\end{minipage}\tabularnewline
\begin{minipage}[t]{0.41\columnwidth}\raggedright
LR annealing strategy\strut
\end{minipage} & \begin{minipage}[t]{0.41\columnwidth}\raggedright
cosine\strut
\end{minipage}\tabularnewline
\begin{minipage}[t]{0.41\columnwidth}\raggedright
LR maximum\strut
\end{minipage} & \begin{minipage}[t]{0.41\columnwidth}\raggedright
2e-4, 1e-4, 1e-3\strut
\end{minipage}\tabularnewline
\begin{minipage}[t]{0.41\columnwidth}\raggedright
LR warmup period\strut
\end{minipage} & \begin{minipage}[t]{0.41\columnwidth}\raggedright
30\%\strut
\end{minipage}\tabularnewline
\begin{minipage}[t]{0.41\columnwidth}\raggedright
LR initial division factor\strut
\end{minipage} & \begin{minipage}[t]{0.41\columnwidth}\raggedright
25\strut
\end{minipage}\tabularnewline
\begin{minipage}[t]{0.41\columnwidth}\raggedright
LR final division factor\strut
\end{minipage} & \begin{minipage}[t]{0.41\columnwidth}\raggedright
1e3\strut
\end{minipage}\tabularnewline
\bottomrule
\end{longtable}

\textsuperscript{1}: To limit the number of total conditions, we set the
LoRA rank (r) equal to the alpha scaling parameter. In other words, if
\texttt{r\ =\ 32}, alpha was also set to 32.

\textsuperscript{2}: When loading the base model with 8-bit integer
quantization, the language modeling head was skipped (i.e.~it was
represented in bfloat16 precision), and the threshold for outlier
handling (\texttt{llm\_int8\_threshold}) was \texttt{6.0}.

Due to the small size of the \texttt{Llama3.2-1B-Instruct} base model,
and the efficiency of fine-tuning with LoRA, training and inference was
performed on an instance with a single NVIDIA L4 GPU.

\clearpage

\hypertarget{a.4-fine-tuning-hyperparameters-closed-weights}{%
\subsection{A.4 Fine-tuning hyperparameters
closed-weights}\label{a.4-fine-tuning-hyperparameters-closed-weights}}

For comparison, we also fine-tuned two closed-weights models from
OpenAI. The configuration options in OpenAI's commercial finetuning
offering are limited. We tried to match all parameters as closely as
possible to those of the open-weight model. The OpenAI finetuning
application does not allow to specify a learning rate, but an ``LR
multiplier''. We did not find an explanation regarding the ``LR
multiplier'' parameter in the official documentation
(https://platform.openai.com/docs). Moreover, the official documentation
does not explain in any detail how finetuning is implemented, but we
assume that OpenAI is using LoRA to efficiently finetune and serve
custom models.

\begin{longtable}[]{@{}ll@{}}
\caption{Hyperparameters for finetuning closed-weights OpenAI
models.}\tabularnewline
\toprule
\begin{minipage}[b]{0.21\columnwidth}\raggedright
Parameter\strut
\end{minipage} & \begin{minipage}[b]{0.67\columnwidth}\raggedright
Set of values\strut
\end{minipage}\tabularnewline
\midrule
\endfirsthead
\toprule
\begin{minipage}[b]{0.21\columnwidth}\raggedright
Parameter\strut
\end{minipage} & \begin{minipage}[b]{0.67\columnwidth}\raggedright
Set of values\strut
\end{minipage}\tabularnewline
\midrule
\endhead
\begin{minipage}[t]{0.21\columnwidth}\raggedright
Method\strut
\end{minipage} & \begin{minipage}[t]{0.67\columnwidth}\raggedright
Supervised\strut
\end{minipage}\tabularnewline
\begin{minipage}[t]{0.21\columnwidth}\raggedright
Base model\strut
\end{minipage} & \begin{minipage}[t]{0.67\columnwidth}\raggedright
gpt-4o-mini-2024-07-18 \& gpt-4.1-mini-2025-04-14\strut
\end{minipage}\tabularnewline
\begin{minipage}[t]{0.21\columnwidth}\raggedright
Seed\strut
\end{minipage} & \begin{minipage}[t]{0.67\columnwidth}\raggedright
713\strut
\end{minipage}\tabularnewline
\begin{minipage}[t]{0.21\columnwidth}\raggedright
Batch size\strut
\end{minipage} & \begin{minipage}[t]{0.67\columnwidth}\raggedright
32\strut
\end{minipage}\tabularnewline
\begin{minipage}[t]{0.21\columnwidth}\raggedright
LR multiplier\strut
\end{minipage} & \begin{minipage}[t]{0.67\columnwidth}\raggedright
1.0\strut
\end{minipage}\tabularnewline
\begin{minipage}[t]{0.21\columnwidth}\raggedright
Epochs\strut
\end{minipage} & \begin{minipage}[t]{0.67\columnwidth}\raggedright
5\strut
\end{minipage}\tabularnewline
\bottomrule
\end{longtable}

\hypertarget{a.5-system-prompts}{%
\subsection{A.5 System prompts}\label{a.5-system-prompts}}

Below is the system prompt used for the base models in Figure 2 and
Table 2. Because this system prompt was used to elicit conversational
answers from these base models without fine-tuning, the system prompt is
fairly specific and verbose:

\begin{quote}
You are a helpful assistant. You answer questions in a natural,
conversational tone, like in a spoken conversation. You give short and
concise answers. Your answers must be one to four sentences long. The
Flesch Reading Ease Score is a metric to assess how difficult a text
passage is to understand. Higher scores indicate text that is easier.
Please formulate your answer such that it would receive a Flesch Reading
Ease Score of above 60.
\end{quote}

In contrast, during finetuning, and when performing inference on the
validation dataset, the following, more concise system prompt was used
(for both the open source and the closed source fine-tuned models):

\begin{quote}
You are a helpful assistant. You answer questions in a natural,
conversational tone, like in a spoken conversation.
\end{quote}

\clearpage

\hypertarget{a.6-detailed-results}{%
\subsection{A.6 Detailed results}\label{a.6-detailed-results}}

\begin{longtable}[]{@{}ll@{}}
\caption{Percentage of conversational responses for two closed-weights
models, and a fine-tuned \texttt{Llama3.2-1B-Instruct} model. Numeric
values corresponding to Figure 2.}\tabularnewline
\toprule
\begin{minipage}[b]{0.33\columnwidth}\raggedright
Model\strut
\end{minipage} & \begin{minipage}[b]{0.41\columnwidth}\raggedright
\% conversational responses\strut
\end{minipage}\tabularnewline
\midrule
\endfirsthead
\toprule
\begin{minipage}[b]{0.33\columnwidth}\raggedright
Model\strut
\end{minipage} & \begin{minipage}[b]{0.41\columnwidth}\raggedright
\% conversational responses\strut
\end{minipage}\tabularnewline
\midrule
\endhead
\begin{minipage}[t]{0.33\columnwidth}\raggedright
gpt-4o-mini base\strut
\end{minipage} & \begin{minipage}[t]{0.41\columnwidth}\raggedright
\begin{verbatim}
                    29.4
\end{verbatim}
\strut
\end{minipage}\tabularnewline
\begin{minipage}[t]{0.33\columnwidth}\raggedright
gpt-4o-mini finetuned\strut
\end{minipage} & \begin{minipage}[t]{0.41\columnwidth}\raggedright
\begin{verbatim}
                    97.8
\end{verbatim}
\strut
\end{minipage}\tabularnewline
\begin{minipage}[t]{0.33\columnwidth}\raggedright
gpt-4.1-mini base\strut
\end{minipage} & \begin{minipage}[t]{0.41\columnwidth}\raggedright
\begin{verbatim}
                    45.6
\end{verbatim}
\strut
\end{minipage}\tabularnewline
\begin{minipage}[t]{0.33\columnwidth}\raggedright
gpt-4.1-mini finetuned\strut
\end{minipage} & \begin{minipage}[t]{0.41\columnwidth}\raggedright
\begin{verbatim}
                    97.6
\end{verbatim}
\strut
\end{minipage}\tabularnewline
\begin{minipage}[t]{0.33\columnwidth}\raggedright
Llama 3.2 1B base\strut
\end{minipage} & \begin{minipage}[t]{0.41\columnwidth}\raggedright
\begin{verbatim}
                    23.1
\end{verbatim}
\strut
\end{minipage}\tabularnewline
\begin{minipage}[t]{0.33\columnwidth}\raggedright
Llama 3.2 1B finetuned\strut
\end{minipage} & \begin{minipage}[t]{0.41\columnwidth}\raggedright
\begin{verbatim}
                    97.4
\end{verbatim}
\strut
\end{minipage}\tabularnewline
\bottomrule
\end{longtable}

\begin{longtable}[]{@{}lllllll@{}}
\caption{Detailed numeric results corresponding to data shown in Figures
2 and 3. For OpenAI models, implementation details are not
known.}\tabularnewline
\toprule
\begin{minipage}[b]{0.13\columnwidth}\raggedright
Model\strut
\end{minipage} & \begin{minipage}[b]{0.10\columnwidth}\raggedright
Data type\strut
\end{minipage} & \begin{minipage}[b]{0.10\columnwidth}\raggedright
LoRA r\strut
\end{minipage} & \begin{minipage}[b]{0.11\columnwidth}\raggedright
Learning rate\strut
\end{minipage} & \begin{minipage}[b]{0.11\columnwidth}\raggedright
Training samples\strut
\end{minipage} & \begin{minipage}[b]{0.12\columnwidth}\raggedright
\% conver- sational responses\strut
\end{minipage} & \begin{minipage}[b]{0.13\columnwidth}\raggedright
Similarity prediction \& correct answer\strut
\end{minipage}\tabularnewline
\midrule
\endfirsthead
\toprule
\begin{minipage}[b]{0.13\columnwidth}\raggedright
Model\strut
\end{minipage} & \begin{minipage}[b]{0.10\columnwidth}\raggedright
Data type\strut
\end{minipage} & \begin{minipage}[b]{0.10\columnwidth}\raggedright
LoRA r\strut
\end{minipage} & \begin{minipage}[b]{0.11\columnwidth}\raggedright
Learning rate\strut
\end{minipage} & \begin{minipage}[b]{0.11\columnwidth}\raggedright
Training samples\strut
\end{minipage} & \begin{minipage}[b]{0.12\columnwidth}\raggedright
\% conver- sational responses\strut
\end{minipage} & \begin{minipage}[b]{0.13\columnwidth}\raggedright
Similarity prediction \& correct answer\strut
\end{minipage}\tabularnewline
\midrule
\endhead
\begin{minipage}[t]{0.13\columnwidth}\raggedright
gpt-4o-mini\strut
\end{minipage} & \begin{minipage}[t]{0.10\columnwidth}\raggedright
unknown\strut
\end{minipage} & \begin{minipage}[t]{0.10\columnwidth}\raggedright
unknown\strut
\end{minipage} & \begin{minipage}[t]{0.11\columnwidth}\raggedright
unknown\strut
\end{minipage} & \begin{minipage}[t]{0.11\columnwidth}\raggedright
\begin{verbatim}
     0
\end{verbatim}
\strut
\end{minipage} & \begin{minipage}[t]{0.12\columnwidth}\raggedright
29.4\strut
\end{minipage} & \begin{minipage}[t]{0.13\columnwidth}\raggedright
0.85\strut
\end{minipage}\tabularnewline
\begin{minipage}[t]{0.13\columnwidth}\raggedright
gpt-4o-mini\strut
\end{minipage} & \begin{minipage}[t]{0.10\columnwidth}\raggedright
unknown\strut
\end{minipage} & \begin{minipage}[t]{0.10\columnwidth}\raggedright
unknown\strut
\end{minipage} & \begin{minipage}[t]{0.11\columnwidth}\raggedright
unknown\strut
\end{minipage} & \begin{minipage}[t]{0.11\columnwidth}\raggedright
\begin{verbatim}
   100
\end{verbatim}
\strut
\end{minipage} & \begin{minipage}[t]{0.12\columnwidth}\raggedright
90.8\strut
\end{minipage} & \begin{minipage}[t]{0.13\columnwidth}\raggedright
0.86\strut
\end{minipage}\tabularnewline
\begin{minipage}[t]{0.13\columnwidth}\raggedright
gpt-4o-mini\strut
\end{minipage} & \begin{minipage}[t]{0.10\columnwidth}\raggedright
unknown\strut
\end{minipage} & \begin{minipage}[t]{0.10\columnwidth}\raggedright
unknown\strut
\end{minipage} & \begin{minipage}[t]{0.11\columnwidth}\raggedright
unknown\strut
\end{minipage} & \begin{minipage}[t]{0.11\columnwidth}\raggedright
\begin{verbatim}
  1000
\end{verbatim}
\strut
\end{minipage} & \begin{minipage}[t]{0.12\columnwidth}\raggedright
96.9\strut
\end{minipage} & \begin{minipage}[t]{0.13\columnwidth}\raggedright
0.87\strut
\end{minipage}\tabularnewline
\bottomrule
\end{longtable}

\hypertarget{a.7-links-to-online-resources}{%
\subsection{A.7 Links to online
resources}\label{a.7-links-to-online-resources}}

The dataset is available at
\url{https://huggingface.co/datasets/restack/conversational-question-answer-wikipedia-v1.0}.

The Llama 3.2 1B model depicted in Figure 2 (i.e.~most successful 1B
parameter Llama model), which was fine-tuned for 5 epochs on the entire
training dataset (9000 question-answer pairs) with r = 64, base model
loaded in 8 bit integer precision, learning rate = 2e-4, is available at
\url{https://huggingface.co/restack/conversational-v1.1-Llama-3.2-1B-Instruct}.

\hypertarget{bibliography}{%
\section*{References}\label{bibliography}}
\addcontentsline{toc}{section}{References}

\hypertarget{refs}{}
\begin{cslreferences}
\leavevmode\hypertarget{ref-bo_quantization_2024}{}%
Bo, Yanan, and Yongqiang Wang. 2024. ``Quantization Avoids Saddle Points
in Distributed Optimization.'' \emph{Proceedings of the National Academy
of Sciences} 121 (17). \url{https://doi.org/10.1073/pnas.2319625121}.

\leavevmode\hypertarget{ref-bondarenko_low-rank_2024}{}%
Bondarenko, Yelysei, Riccardo Del Chiaro, and Markus Nagel. 2024.
``Low-Rank Quantization-Aware Training for LLMs.'' arXiv.
\url{https://doi.org/10.48550/arXiv.2406.06385}.

\leavevmode\hypertarget{ref-brown_language_2020}{}%
Brown, Tom B., Benjamin Mann, Nick Ryder, Melanie Subbiah, Jared Kaplan,
Prafulla Dhariwal, Arvind Neelakantan, et al. 2020. ``Language Models
Are Few-Shot Learners.'' arXiv.
\url{https://doi.org/10.48550/arXiv.2005.14165}.

\leavevmode\hypertarget{ref-dettmers_llmint8_2022}{}%
Dettmers, Tim, Mike Lewis, Younes Belkada, and Luke Zettlemoyer. 2022.
``LLM.Int8(): 8-Bit Matrix Multiplication for Transformers at Scale.''
arXiv. \url{https://doi.org/10.48550/arXiv.2208.07339}.

\leavevmode\hypertarget{ref-grattafiori_llama_2024}{}%
Grattafiori, Aaron, Abhimanyu Dubey, Abhinav Jauhri, Abhinav Pandey,
Abhishek Kadian, Ahmad Al-Dahle, Aiesha Letman, et al. 2024. ``The Llama
3 Herd of Models.'' arXiv.
\url{https://doi.org/10.48550/arXiv.2407.21783}.

\leavevmode\hypertarget{ref-han_parameter-efficient_2024}{}%
Han, Zeyu, Chao Gao, Jinyang Liu, Jeff Zhang, and Sai Qian Zhang. 2024.
``Parameter-Efficient Fine-Tuning for Large Models: A Comprehensive
Survey.'' arXiv. \url{https://doi.org/10.48550/arXiv.2403.14608}.

\leavevmode\hypertarget{ref-hu_lora_2021}{}%
Hu, Edward J., Yelong Shen, Phillip Wallis, Zeyuan Allen-Zhu, Yuanzhi
Li, Shean Wang, Lu Wang, and Weizhu Chen. 2021. ``LoRA: Low-Rank
Adaptation of Large Language Models.'' arXiv.
\url{https://doi.org/10.48550/arXiv.2106.09685}.

\leavevmode\hypertarget{ref-kincaid_derivation_1975}{}%
Kincaid, J., Robert Fishburne, Richard Rogers, and Brad Chissom. 1975.
``Derivation of New Readability Formulas (Automated Readability Index,
Fog Count and Flesch Reading Ease Formula) for Navy Enlisted
Personnel.'' \emph{Institute for Simulation and Training}, January.
\url{https://stars.library.ucf.edu/istlibrary/56}.

\leavevmode\hypertarget{ref-kusano_are_2024}{}%
Kusano, Genki, Kosuke Akimoto, and Kunihiro Takeoka. 2024. ``Are Longer
Prompts Always Better? Prompt Selection in Large Language Models for
Recommendation Systems.'' arXiv.
\url{https://doi.org/10.48550/ARXIV.2412.14454}.

\leavevmode\hypertarget{ref-lee_predictive_2025}{}%
Lee, Jae Yong, Sungmin Kang, and Shin Yoo. 2025. ``Predictive Prompt
Analysis.'' arXiv. \url{https://doi.org/10.48550/arXiv.2501.18883}.

\leavevmode\hypertarget{ref-li_towards_2023}{}%
Li, Zehan, Xin Zhang, Yanzhao Zhang, Dingkun Long, Pengjun Xie, and
Meishan Zhang. 2023. ``Towards General Text Embeddings with Multi-Stage
Contrastive Learning.'' arXiv.
\url{https://doi.org/10.48550/arXiv.2308.03281}.

\leavevmode\hypertarget{ref-loshchilov_decoupled_2019}{}%
Loshchilov, Ilya, and Frank Hutter. 2019. ``Decoupled Weight Decay
Regularization.'' arXiv.
\url{https://doi.org/10.48550/arXiv.1711.05101}.

\leavevmode\hypertarget{ref-lu_fantastically_2022}{}%
Lu, Yao, Max Bartolo, Alastair Moore, Sebastian Riedel, and Pontus
Stenetorp. 2022. ``Fantastically Ordered Prompts and Where to Find Them:
Overcoming Few-Shot Prompt Order Sensitivity.'' arXiv.
\url{https://doi.org/10.48550/arXiv.2104.08786}.

\leavevmode\hypertarget{ref-luo_empirical_2025}{}%
Luo, Yun, Zhen Yang, Fandong Meng, Yafu Li, Jie Zhou, and Yue Zhang.
2025. ``An Empirical Study of Catastrophic Forgetting in Large Language
Models During Continual Fine-Tuning.'' arXiv.
\url{https://doi.org/10.48550/arXiv.2308.08747}.

\leavevmode\hypertarget{ref-min_rethinking_2022}{}%
Min, Sewon, Xinxi Lyu, Ari Holtzman, Mikel Artetxe, Mike Lewis, Hannaneh
Hajishirzi, and Luke Zettlemoyer. 2022. ``Rethinking the Role of
Demonstrations: What Makes in-Context Learning Work?'' arXiv.
\url{https://doi.org/10.48550/arXiv.2202.12837}.

\leavevmode\hypertarget{ref-polo_efficient_2024}{}%
Polo, Felipe Maia, Ronald Xu, Lucas Weber, Mírian Silva, Onkar Bhardwaj,
Leshem Choshen, Allysson Flavio Melo de Oliveira, Yuekai Sun, and
Mikhail Yurochkin. 2024. ``Efficient Multi-Prompt Evaluation of LLMs.''
arXiv. \url{https://doi.org/10.48550/arXiv.2405.17202}.

\leavevmode\hypertarget{ref-radford_robust_2022}{}%
Radford, Alec, Jong Wook Kim, Tao Xu, Greg Brockman, Christine McLeavey,
and Ilya Sutskever. 2022. ``Robust Speech Recognition via Large-Scale
Weak Supervision.'' arXiv.
\url{https://doi.org/10.48550/arXiv.2212.04356}.

\leavevmode\hypertarget{ref-schuhmann_wikipedia-en-chunks_2024}{}%
Schuhmann, Christoph. 2024. ``Wikipedia-En-Chunks.''
\url{https://huggingface.co/datasets/ChristophSchuhmann/wikipedia-en-chunks}.

\leavevmode\hypertarget{ref-warner_smarter_2024}{}%
Warner, Benjamin, Antoine Chaffin, Benjamin Clavié, Orion Weller, Oskar
Hallström, Said Taghadouini, Alexis Gallagher, et al. 2024. ``Smarter,
Better, Faster, Longer: A Modern Bidirectional Encoder for Fast, Memory
Efficient, and Long Context Finetuning and Inference.'' arXiv.
\url{https://doi.org/10.48550/arXiv.2412.13663}.

\leavevmode\hypertarget{ref-wen_benchmarking_2024}{}%
Wen, Bosi, Pei Ke, Xiaotao Gu, Lindong Wu, Hao Huang, Jinfeng Zhou,
Wenchuang Li, et al. 2024. ``Benchmarking Complex Instruction-Following
with Multiple Constraints Composition.'' arXiv.
\url{https://doi.org/10.48550/arXiv.2407.03978}.

\leavevmode\hypertarget{ref-xu_parameter-efficient_2023}{}%
Xu, Lingling, Haoran Xie, Si-Zhao Joe Qin, Xiaohui Tao, and Fu Lee Wang.
2023. ``Parameter-Efficient Fine-Tuning Methods for Pretrained Language
Models: A Critical Review and Assessment.'' arXiv.
\url{https://doi.org/10.48550/arXiv.2312.12148}.

\leavevmode\hypertarget{ref-zhao_calibrate_2021}{}%
Zhao, Tony Z., Eric Wallace, Shi Feng, Dan Klein, and Sameer Singh.
2021. ``Calibrate Before Use: Improving Few-Shot Performance of Language
Models.'' arXiv. \url{https://doi.org/10.48550/arXiv.2102.09690}.
\end{cslreferences}

\end{document}